\documentclass[10pt,twocolumn,letterpaper]{article}

\usepackage{iccv}
\usepackage{times}
\usepackage{epsfig}
\usepackage{graphicx}
\usepackage{amsmath}
\usepackage{amssymb}

\usepackage[ruled,vlined]{algorithm2e}
\DeclareMathOperator*{\argmax}{arg\,max}
\usepackage{subcaption}
\usepackage{multirow}

\usepackage[breaklinks=true,bookmarks=false]{hyperref}

\iccvfinalcopy % *** Uncomment this line for the final submission

\def\iccvPaperID{****} % *** Enter the ICCV Paper ID here

\ificcvfinal\fi

\begin{document}

%%%%%%%%% TITLE
\title{Task-agnostic Out-of-Distribution Detection Using Kernel Density Estimation}

\author{Ertunc Erdil, Krishna Chaitanya, Neerav Karani, Ender Konukoglu \\
Computer Vision Lab, ETH Zurich \\
Sternwartstrasse 7, Zurich 8092, Switzerland \\
{\tt\small ertunc.erdil@vision.ee.ethz.ch}
}

\maketitle
\ificcvfinal\thispagestyle{empty}\fi

%%%%%%%%% ABSTRACT
\begin{abstract}
In the recent years, researchers proposed a number of successful methods to perform out-of-distribution (OOD) detection in deep neural networks (DNNs). 
So far the scope of the highly accurate methods has been limited to image level classification tasks. 
However, attempts for generally applicable methods beyond classification did not attain similar performance. 
In this paper, we address this limitation by proposing a simple yet effective task-agnostic OOD detection method.
We estimate the probability density functions (pdfs) of intermediate features of a pre-trained DNN by performing kernel density estimation (KDE) on the training dataset.
As direct application of KDE to feature maps is hindered by their high dimensionality, we use a set of lower-dimensional marginalized KDE models instead of a single high-dimensional one.
At test time, we evaluate the pdfs on a test sample and produce a confidence score that indicates the sample is OOD.
The use of KDE eliminates the need for making simplifying assumptions about the underlying feature pdfs and makes the proposed method task-agnostic.
We perform extensive experiments on classification tasks using benchmark datasets for OOD detection. 
Additionally, we perform experiments on medical image segmentation tasks using brain MRI datasets.
The results demonstrate that the proposed method consistently achieves high OOD detection performance in both classification and segmentation tasks and improves state-of-the-art in almost all cases.
Code is available at \url{https://github.com/eerdil/task_agnostic_ood}
\end{abstract}

% ======================================================
% INTRODUCTION
% ======================================================
\section{Introduction}
Deep neural networks (DNNs) can perform predictions on test images with very high accuracy when the training and testing data come from the same distribution. 
However, the prediction accuracy decreases rapidly when the test image is sampled from a different distribution than the training one \cite{karani2021test, wang2020fully}. 
Furthermore, in such cases, DNNs can make erroneous predictions with very high confidence \cite{guo2017calibration}. 
This creates a major obstacle when deploying DNNs for real applications, especially for the ones with a low tolerance for error, such as autonomous driving and medical diagnosis.
Therefore, it is crucial to improve the robustness of DNN-based methods and prevent them from making big mistakes \cite{amodei2016concrete}. 

\begin{figure}[t]
    \centering
    \begin{tabular}{ccc}
     Input &  Prediction & Ground Truth \\
    \includegraphics[scale = 0.26]{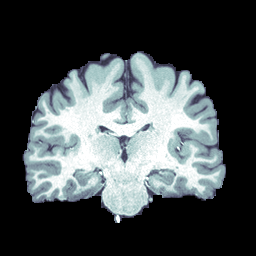} &
    \includegraphics[scale = 0.26]{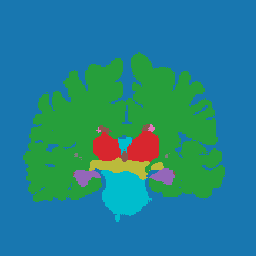} &
    \includegraphics[scale = 0.26]{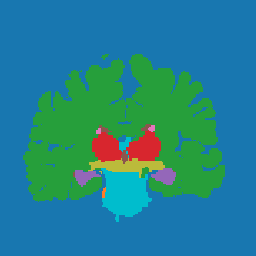} \\
    \includegraphics[scale = 0.26]{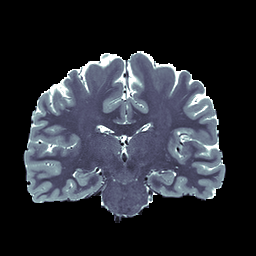} &
    \includegraphics[scale = 0.26]{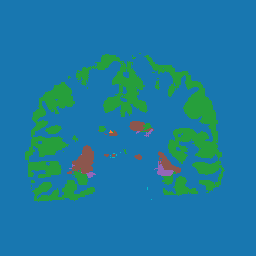} &
    \includegraphics[scale = 0.26]{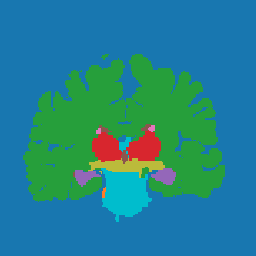}
    \end{tabular}
    \caption{A visual example demonstrating the importance of OOD detection on a segmentation task. A network trained on T1w images works well on a T1w test image (first row) while it produces a poor segmentation on the T2w image of the same patient (second row).}
    \label{fig:visual_example}
\end{figure}
Recently, to improve the robustness of DNNs,substantial advances have been made for OOD detection in DNNs trained for image level classification tasks \cite{hendrycks17baseline, lee2018simple, liu2020energy, hsu2020generalized}.
Although, OOD detection is equally crucial for non-classification tasks (e.g. segmentation), so far, attempts for developing more generic OOD detection methods did not attain similar performance \cite{nalisnick2018deep}.
In Figure \ref{fig:visual_example}, we show a visual example in a medical image segmentation task to emphasize the importance of OOD detection in a non-classification task.
The example demonstrates that a DNN trained on T1w brain images (in-distribution (InD)) produces poor segmentation results for T2w images (OOD) of the same patient.
Let us imagine an automated clinical decision system that makes the diagnosis based on the segmentations.
In that case, the poor segmentation of the OOD T2w image could lead to a wrong diagnosis which can have severe consequences.
An OOD detection method within this pipeline can play a key role in preventing such mistakes.

% ======================================================
% RELATED WORK
% ======================================================
\subsection{Related Work}
\label{sec:related_work}
% We describe the OOD detection methods in the literature in three categories.

\vspace{0.1cm} \noindent \textbf{Methods that use predicted class probabilities}:
Hendrycks et al.~\cite{hendrycks17baseline} proposed a baseline OOD detection method that uses maximum predicted class probability in a classification DNN as a confidence score that the sample is OOD. 
% The performance of this method diminishes when DNNs make overconfident predictions on OOD images, which is quite common \cite{guo2017calibration}.
ODIN~\cite{liang2017enhancing} extends the baseline by applying an adversarial perturbation to the input image (referred to as input pre-processing) and temperature scaling before softmax to to increase the difference between the prediction probabilities of InD and OOD samples. 
% Setting the parameters for perturbation strength and temperature scaling are crucial for ODIN. These parameters are determined using a validation set that contains examples from the target OOD dataset, which is usually not available in a real application.
The method proposed by Sastry et al. \cite{sastry2019detecting} take the prediction of an input image and compare the intermediate features of the image with the ones extracted from other images of the predicted class using gram matrices.
The work of Hsu et al.~\cite{hsu2020generalized} further extended ODIN, referred to as Generalized ODIN (G-ODIN), by introducing an additional output that indicates whether the input sample belongs to InD or OOD. The penultimate layer of a DNN is decomposed into two branches to model the conditional distribution of this indicator variable and its joint distribution with the class label. The conditional probability of the indicator variable is used as the confidence score while the ratio of the joint probability and the conditional probability serve as the final class prediction for an image after applying input processing.

\vspace{0.1cm} \noindent \textbf{Methods that propose training strategies}:
DeVries and Taylor~\cite{devries2018learning} introduced a confidence estimation network branch, and proposed to use the confidence estimates obtained from this branch in order to train the softmax probabilities of the classification network. 
Lee et al.~\cite{lee2018training} presented a training method for classifier networks so that they become less confident for OOD examples. They introduced two loss terms in addition to the cross-entropy. The first one encourages the network to become less confident for the OOD examples whereas the second one generates the most optimal OOD examples for the first one. 
The method of Vyas et al.~\cite{vyas2018out} proposed to use an ensemble of classifiers for OOD detection, where each classifier is trained by leaving-out a different class from the InD training set. OOD detection is then performed based on the ensemble of softmax probabilities of each classifier, after applying temperature scaling and input pre-processing as in ODIN. 
Hendrycks et al. \cite{hendrycks2018deep} proposed a method called outlier exposure which exploits existing very large datasets that are known to be OOD during training. 
Yu et al. \cite{yu2019unsupervised} proposed a DNN with two classification heads, where one aims to minimize classification loss and the other aims to maximize the discrepancy between the two classifiers. The method, named as MCD, uses a subset of OOD samples along with the InD samples in the discrepancy loss. At test time, the samples with higher discrepancy are labeled as OOD. 
Very recently, Liu et al. \cite{liu2020energy} proposed an energy-based method (EBM) which interprets softmax probabilities as energy scores and use for OOD detection. 

\vspace{0.1cm} \noindent \textbf{Methods that perform density estimation}:
Lee et al.~\cite{lee2018simple} proposed a method named as Mahalanobis, which models the class conditional pdfs of the features at intermediate layers of a DNN with Gaussian densities for InD samples. The parameters of each class conditional Gaussian are estimated by computing the empirical mean and co-variance using InD training samples belonging to that class. At test time, ODIN-style input pre-processing is applied before evaluating the estimated densities to obtain a confidence score, which is expected to be higher for InD samples and lower for OOD samples.

Despite their successful performance, most of the aforementioned methods are designed for OOD detection for classification tasks and their extension to non-classification tasks usually is not trivial.
Task-agnostic networks that do not share the same drawback have also been proposed. 
Hendryks et al.~\cite{hendrycks2019using} proposed a self-supervised learning (SSL) based OOD detection method. The method trains an auxiliary rotation network, which predicts angle of rotation in discrete categories, on the InD dataset and computes a confidence score for a test image as the maximum of the softmax activation, expecting higher activations for InD samples compared to OOD samples. 
Kim et al. \cite{kim2020rapp} proposed a method, referred to as RaPP, that is based on the observation that in an autoencoder the internal feature representations of an input image and its reconstructed version are very similar for InD samples and the similarity decreases for OOD samples that are not used for the training of the autoencoder. RaPP defines a confidence score based on this observation for OOD detection. 
Venkatakrishnan et al. \cite{venkatakrishnan2020self} combine the ideas in SSL and RaPP, and propose a method (Multitask\_SSL) by jointly training a network for both rotation prediction and reconstruction tasks for OOD detecion in brain images. 
As both SSL \cite{hendrycks2019using}, RaPP \cite{kim2020rapp}, and Multitask\_SSL\cite{venkatakrishnan2020self} operate on auxiliary networks that are detached from the main network, they are task-agnostic and therefore can be applied to both classification and non-classification tasks.

% ======================================================
% CONTRIBUTION
% ======================================================
\subsection{Contribution}
In this paper, we propose a simple yet effective task-agnostic OOD detection method. In the proposed method, we estimate feature pdfs of each channel in a DNN using KDE and InD training images. We evaluate the pdfs using a new test sample and obtain a confidence score for each channel. We combine all the scores into a final confidence score using a logistic regression model, which we train using channel-wise confidence scores of training images as InD samples and their adversarially perturbed versions as OOD samples.

We take our motivation from Mahalanobis~\cite{lee2018simple} for developing the proposed method but extend it in multiple ways crucial for building a task-agnostic method that achieves improved detection accuracy. (1) Mahalanobis estimates class conditional densities while the distribution approximation in the proposed method is not conditioned on the class, making it task-agnostic. (2) Direct use of Mahalanobis in the task-agnostic setting is feasible using unconditioned Gaussian distributions to approximate layer-wise feature distributions of InD samples. However, the Gaussian assumption may be too restrictive to model unconditioned feature densities and lead to lower accuracy when the assumption does not hold. Using a nonparametric density estimation method KDEs, we extend the flexibility of the density approximation in the proposed method. (3) Layer-wise approximation is prone to the curse-of-dimensionality. Even though Mahalanobis takes channel-wise mean to reduce the dimension of a channel from $C \times H \times W$ to $C \times 1$ before density estimation, the resulting vector can still be high dimensional in modern architectures. We approximate 1D channel-wise distributions in the proposed method, which are simpler to estimate. This approach ignores dependencies between channels of a layer in the density estimation part but takes them into account in the logistic regression model that combines channel-wise scores. 

The use of KDEs for OOD detection is not new. The first use dates back to 1994 when Bishop \cite{bishop1994novelty} applied KDE in the input space. In that work, the input was only 12-dimensional and application of KDE was feasible. In modern architectures, the input dimensions are often much larger thereby, making the direct application of Bishop's method infeasible. The application of a modified version of the method in high dimensional spaces is still possible by applying KDE over the distances between the test image and the training images~\cite{kim2007nonparametric, cremers2006kernel}. Bishop's method as well as its modified version differs from the proposed method. In our method, we use multiple channel-wise KDEs and aggregate results

Using KDE and estimating channel-wise pdfs are conceptually simple extensions that are very effective and yield substantial gains. We performed extensive comparisons on DNNs trained for classification and segmentation tasks. In the classification experiments, we use the common benchmark that contains 2 different classification networks trained on CIFAR-10 and CIFAR-100 datasets \cite{krizhevsky2009learning} and 6 other OOD datasets. We compare the proposed method with 6 methods in the literature most of which are either very recent or common baselines used in the literature. In the segmentation experiments, we use datasets for brain MRI segmentation and compare with 5 methods. In total, we compare with 10 different OOD detection methods to correctly position the proposed method within the current literature.

% ======================================================
% METHOD
% ======================================================
\section{Method}
Let us denote a set of training images with $X_{tr} = \{x_1, x_2, \dots x_M\} \sim P_{in}$ and corresponding labels with $y_{tr} = \{y_1, y_2, \dots, y_M\}$, where $P_{in}$ denotes the InD. Let us also denote a DNN with $f$, trained using $(X_{tr}, y_{tr})$. $f$ is more likely to perform good predictions on a test image $x_{test}$ if $x_{test} \sim P_{in}$ and incorrect predictions if $x_{test} \sim P_{out}$, where $P_{out} \neq P_{in}$. In this section, we present the proposed task-agnostic KDE-based approach that identifies test images sampled from $P_{out}$. The application of the proposed model at test time is summarized in Algorithm \ref{alg:method_general}.

% ============================
% Algorithm
% ============================
\begin{algorithm}
\SetAlgoLined
\KwIn{Test Image: $x$

Random subset of training images: $\hat{X}_{tr} = \{x_{u_{1}}, x_{u_{2}}, \dots, x_{u_{N}}\}$

Weights of logistic regression classifier: $\alpha_{lc}$ 

Set of kernel sizes: $\boldsymbol \sigma$ 
}
\KwOut{Confidence score of $x$: $\mathcal{M}_x$.}
\ForEach {$l \in L$}{
% \For{$l\gets1$ \KwTo L}{
    \For{$c\gets1$ \KwTo $C_l$}{
        Perform KDE at channel $c$ of the feature map at layer $l$: $\hat{p}_{lc}(x) = \frac{1}{N} \sum_{i = 1}^N
        \mathcal{K}(f_{lc}(x) - f_{lc}(x_{u_{i}}); \sigma_{lc})$
    }
 }
 \Return $\mathcal{M}_x = \sum_{l = 1}^{L} \sum_{c = 1}^{C_l} \alpha_{lc} \hat{p}_{lc}(x)$
 \caption{The proposed method\label{alg:method_general}}
\end{algorithm}

% ============================
% Computation of KDE confidence scores
% ============================
\subsection{Computing the confidence scores}
\label{sec:confidence}
The main output of the proposed method is a confidence score that indicates how likely a given sample belongs to OOD for the given DNN $f$. In this section, we describe how we compute this score.
Let us assume that $f$ consists of $L$ layers and the feature map in a layer $l$ for a given image $x$ is denoted as $f_l(x)$ and it has dimensions $C_l \times H_l \times W_l$, where $C_l$, $H_l$, and $W_l$ are the number of channels, height, and width of the feature map, respectively.
We take the channel-wise mean of the feature map and reduce the dimensionality to $C_l \times 1$, as also done in \cite{lee2018simple}. 
We denote the resulting $C_l$-dimensional feature vector by $f'_{l}(x)$. 
We then estimate the marginal feature pdfs \emph{for each channel} $c$ using KDE:
\begin{equation}
    p_{lc}(x) \approx \hat{p}_{lc}(x) = \frac{1}{M} \sum_{i = 1}^M \mathcal{K}(f'_{lc}(x) - f'_{lc}(x_i); \sigma_{lc})
    \label{eq:kde_1}
\end{equation}
where $p_{lc}$ is the true marginal pdf of the features $f'_{lc}$ in channel $c$ of layer $l$, $\hat{p}_{lc}$ is the estimate of the pdf, and $\mathcal{K}(u,v;\sigma_{lc})=e^{-(u-v)^2/\sigma_{lc}^2}$ is a 1D squared exponential kernel with $\sigma_{lc}$ being the kernel size.
We describe the procedure to set the kernel size $\sigma_{lc}$ in detail later in Section~\ref{sec:kernel_size}. When using KDE, we use samples $x_i\in X_{tr}$ and thus model InD channel-wise pdfs with $\hat{p}_{lc}$. For a given sample $x$, $\hat{p}_{lc}(x)$ is the confidence score of channel $c$ in layer $l$. 

\vspace{0.1cm} \noindent The advantage of estimating channel-wise pdfs over estimating layer-wise pdfs, as was done in~\cite{lee2018simple}, is performing density estimation in 1D space instead of $C_l-$D space. Typically, $C_l$ can be very large in modern networks, and density estimation becomes less accurate in high-dimensions~\cite{scott2015multivariate}, channel-wise estimation avoids this.

\vspace{0.1cm} \noindent In order to evaluate $p_{lc}$ in Eq.\ (\ref{eq:kde_1}) for a new sample, ideally we need to store all the InD training images in $X_{tr}$. In real-world applications where $M$ is very large, storing the entire $X_{tr}$ may not be feasible and the summation over $M$ images in Eq.\ (\ref{eq:kde_1}) can take very long. Improving the computational and memory efficiency of KDE-based methods are possible by defining an unbiased estimator \cite{erdil2019pseudo}, which simply uses a random subset of $X_{tr}$ such that 
\[
\hat{X}_{tr} = \{x_{u_{1}}, x_{u_{2}}, \dots, x_{u_{N}}\} \subset X_{tr}
\]
where $\{u_1, u_2, \dots, u_N\} \subset \{1, 2, \dots, M\}$ is a random subset of indices generated by sampling from a Uniform density, $\mathcal{U}(1, M)$, without replacement and $N << M$. Using the random subset, we replace the summation in Eq.\ (\ref{eq:kde_1}) with the computationally more efficient unbiased estimator
\begin{equation}
    p_{lc}(x) \approx \hat{p}_{lc}(x) = \frac{1}{N} \sum_{i = 1}^N \mathcal{K}(f'_{lc}(x) -  f'_{lc}(x_{u_i}); \sigma_{lc}).
    \label{eq:kde_2}
\end{equation}
In our experiments, we set $N = 5000$. In the supplementary material, we demonstrate results with different choices of $N$.

Estimating marginal pdfs using Eq.\ (\ref{eq:kde_2}) does not model dependencies between channels. In the proposed method, we take into account such dependencies and compute the final confidence score using a logistic regression classifier:
\begin{equation}
\mathcal{M}_x = \sum_{l = 1}^{L} \sum_{c = 1}^{C_l} \alpha_{lc} \hat{p}_{lc}(x)
\label{eq:final_score}
\end{equation}
where $\alpha_{lc}$ are the weights that are learned as described next.

% ============================
% Training logistic regression
% ============================
\subsection{Learning logistic regression weights $\alpha_{lc}$}
\label{sec:regression}
The role of the logistic regression model is to distinguish between InD and OOD samples given the channel-wise confidence scores. 
Training for the weights $\alpha_{lc}$ requires having access to both InD and OOD images. Although the InD images, $X_{tr}$, are already available, it is difficult to capture all possible images in $P_{out}$. Lee et al. \cite{lee2018simple} propose using adversarial examples obtained by FGSM \cite{goodfellow2014explaining} as samples from $P_{out}$ for hyperparameter tuning. We use adversarial examples as OOD samples to train logistic regression in the proposed method. After obtaining OOD samples by applying adversarial perturbation to the images in $X_{tr}$ using FGSM, the logistic regression classifier is trained by using the confidence scores $\hat{p}_{lc}$ as inputs, and the output labels are provided as positive for InD images and negative for the OOD ones. Note that FGSM can work with any type of label that is for the task, e.g., image-level label for classification, ground truth mask for segmentation and so on. Therefore, using FGSM does not affect the task-agnostic nature of the proposed method. We present further details on the FGSM method in the supplementary materials for completeness. 

% ============================
% Determining kernel size
% ============================
\subsection{Determining size of KDE kernel $\sigma_{lc}$}
\label{sec:kernel_size}
Kernel size, $\sigma$, is a crucial parameter of KDE since it significantly affects the shape of the estimated density. Setting $\sigma$ to a large value leads to very smooth pdfs, reducing the likelihood of the observed samples as well as other samples from the same distribution. Setting $\sigma$ to a very small value leads to very peaky distributions only attributing high probability to observed samples, and assigning very low probability to unobserved samples even from the same distribution. Therefore, finding an optimal $\sigma$ value is quite important in order to capture the underlying pdf of the data \cite{scott2015multivariate}. 

\vspace{0.1cm} \noindent In the proposed method, as given in Eq. (\ref{eq:kde_2}), we use a different $\sigma_{lc}$ for each channel $c$ and layer $l$. To compute $\sigma_{lc}$, we use the $k$-nearest neighbor method ($k$NN) \cite{silverman1986density}. Specifically, we compute the distance between $f'_{lc}(x_{u_{i}})$ and $f_{lc}(x_{u_{j}})$ for all $i, j \in [1, N]$, $i \neq j$ and set $\sigma_{lc}$ as the $k^{th}$ smallest value. We denote the set of all $\sigma_{lc}$ values by $\boldsymbol \sigma$.

\vspace{0.1cm} \noindent A parameter of the $k$NN method is the $k$ itself. We automatically select the most appropriate $k$ value from a set of candidate values denoted by $\mathbf{k}$. To achieve this, we split a validation set $X_{val}$ from $X_{tr}$ by taking the samples that are not used in KDE. Then, we apply adversarial perturbation to the images in $X_{val}$ and obtain $X_{val}^{adv}$ as OOD examples. We select the $k \in \mathbf{k}$ that maximizes the difference between the InD and OOD datasets.
\begin{equation}
    k_{lc} = \argmax_{k \in \mathbf{k}} \sum_{x \in X_{val}} \hat{p}_{lc}(x) - \sum_{x' \in X_{val}^{adv}} \hat{p}_{lc}(x')
\end{equation}
where $k_{lc}$ indicates the optimum $k$ value in layer $l$. In our experiments, we choose $k_l$ from the candidate set $\mathbf{k} = \{1, 2, 5, 10, 15, 20, 50\}$ using the described method. 

% ======================================
% EXPERIMENTS
% ======================================
\section{Experiments and results}
We evaluate the performance of the proposed approach on DNNs trained for classification and segmentation tasks. The proposed method is implemented in PyTorch and we run all experiments on a Nvidia GeForce Titan X GPU with 12GB memory.

% ======================================
% DATASETS AND NETWORK ARCHITECTURES
% ======================================
\subsection{Datasets and network architectures}
\label{sec:dataset}
In the classification experiments, we use two different ResNet architectures trained on CIFAR-10 and CIFAR-100 \cite{krizhevsky2009learning} datasets that contain images from $10$ and $100$ classes, respectively. Both datasets contain $50000$ training and $10000$ test RGB color images of size $32\times32$.
The pretrained models are used as common benchmarks in the literature and are available at https://github.com/pokaxpoka/deep\_Mahalanobis\_detector. 

\vspace{0.1cm} \noindent We use 6 common benchmark datasets as OOD. SVHN \cite{netzer2011reading} contains $26032$ images of house numbers in Google street-view images. TinyImageNet (TIN) dataset consists of $10000$ $32 \times 32$ RGB test images \cite{deng2009imagenet}. LSUN dataset contain $10000$ RGB test images with size $32 \times 32$ \cite{yu2015lsun}. iSUN is a subset of SUN dataset \cite{xu2015turkergaze} that contains 8925 RGB images resized to $32\times32$. Gaussian and Uniform datasets contain 10000 noise images generated from a Gaussian distribution with zero mean and unit variance, and a Uniform distribution, respectively. SVHN dataset is available in Pytorch and TIN, LSUN, and iSUN are available at https://github.com/facebookresearch/odin.

In the segmentation experiments, we use images from 2 publicly available datasets for brain segmentation: Human Connectome Project (HCP) \cite{van2013wu} and Autism Brain Imaging Data Exchange (ABIDE) \cite{di2014autism}. HCP dataset contains both T1w and T2w images for each subject, while ABIDE dataset consists of T1w image from different imaging sites. HCP\_T1w and HCP\_T2w datasets contain images from 47 patients and we split 21 for training, 5 for validation, and 21 for testing. There are T1w images from 37 patients in both ABIDE\_Caltech\_T1w and ABIDE\_Stanford\_T1w datasets and we split 11, 5, 21 images for train, validation and test.

\vspace{0.1cm} \noindent Using HCP and ABIDE datasets, we design 2 different experiments to evaluate OOD detection performance on segmentation task. In the first experiment, we train a UNet \cite{ronneberger2015u} architecture on ABIDE\_Caltech\_T1w images and use ABIDE\_Stanford\_T1w, HCP\_T1w, and HCP\_T2w images as OOD. In the second experiment, we train the UNet on HCP\_T1w image and use ABIDE\_Caltech\_T1w, ABIDE\_Stanford\_T1w, and HCP\_T2w as OOD. We choose UNet as the network architecture since it is the most common choice for medical image segmentation \cite{ronneberger2015u,chaitanya2020contrastive, kohl2018probabilistic}. In both experiments, we segment the following 15 labels: background, cerebellum gray matter, cerebellum white matter, cerebral gray matter, cerebral white matter, thalamus, hippocampus, amygdala, ventricles, caudate, putamen, pallidum, ventral DC, CSF and brain stem. 
% The pre-trained segmentation models will be made available.

% ======================================
% TABLES CLASSIFICATION
% ======================================
\begin{table*}[!]
\centering
\begin{tabular}{ccc}
\hline
OOD     &   FPR at 95\% TPR $\downarrow$    &   AUROC $\uparrow$ \\ \cline{2-3}
~ & \multicolumn{2}{c}{Baseline / ODIN / Mahalanobis / MCD / G-ODIN / EBM / \textbf{Proposed}} \\ \hline
SVHN    & 25.77 / 16.65 / 8.37  / 60.61 / 10.50 / 6.86  / \textbf{6.49}
        & 89.88 / 95.42 / 98.12 / 72.86 / 97.80 / 98.19 / \textbf{98.48} \\ \hline
        
TIN     & 28.37 / 11.24 / 18.89 / 40.44 / 18.60 / 35.88 / \textbf{8.41}
        & 90.53 / 96.78 / 96.73 / 89.75 / 96.10 / 86.21 / \textbf{98.31} \\ \hline

LSUN    & 28.31 / 10.30 / 19.61 / 34.46 / 9.10  / 21.62 / \textbf{3.80}
        & 91.09 / 97.06 / 96.77 / 91.15 / 98.00 / 92.50 / \textbf{99.01} \\ \hline

iSUN    & 28.02 / 12.37 / 22.46 / 37.72 / 11.20 / 22.52 / \textbf{7.31}
        & 91.01 / 96.03 / 96.34 / 89.89 / 97.60 / 92.03 / \textbf{98.55} \\ \hline

Gaussian& 6.44  / 2.69 / \textbf{0.0}   / 4.21  / \textbf{0.0}   / 0.13  / \textbf{0.0}
        & 97.11 / 98.45 / \textbf{100.0} / 97.14 / \textbf{100.0} / 99.96 / \textbf{100.0} \\ \hline

Uniform & 9.24  / 4.16 / \textbf{0.0}   / 13.17 / \textbf{0.0}   / \textbf{0.0}   / \textbf{0.0}
        & 96.04 / 97.78 / \textbf{100.0} / 92.69 / \textbf{100.0} / \textbf{100.0} / \textbf{100.0} \\ \hline
\end{tabular}
\caption{Quantitative results on distinguishing test set of CIFAR-10 InD dataset from several OOD datasets. $\uparrow$ indicates larger value is better and $\downarrow$ indicates lower value is better. \textit{All values are percentages.}
\label{tab:cifar10}}
\end{table*}
\begin{table*}[!]
\centering
\begin{tabular}{ccc}
\hline
OOD     &   FPR at 95\% TPR $\downarrow$    &   AUROC $\uparrow$ \\ \cline{2-3}
~ & \multicolumn{2}{c}{Baseline / ODIN / Mahalanobis / MCD / G-ODIN / EBM / \textbf{Proposed}} \\ \hline
SVHN    & 55.73 / 24.76 / \textbf{15.53} / 73.33 / 44.90 / 45.49 / 17.46
        & 79.34 / 92.13 / \textbf{97.01} / 64.92 / 93.20 / 88.93 / 95.44 \\ \hline

TIN     & 58.97 / 33.74 / 24.33 / 56.95 / 23.50 / 70.04 / \textbf{7.64}
        & 77.01 / 88.32 / 95.04 / 85.53 / 95.90 / 75.11 / \textbf{98.38} \\ \hline

LSUN    & 64.71 / 37.09 / 28.68 / 58.40 / 23.20 / 67.99 / \textbf{3.73}
        & 75.58 / 87.70 / 94.66 / 84.97 / 96.10 / 76.45 / \textbf{99.13} \\ \hline

iSUN    & 63.26 / 38.21 / 29.46 / 64.32 / 24.70 / 70.11 / \textbf{6.07}
        & 75.68 / 86.73 / 94.02 / 83.46 / 95.70 / 76.57 / \textbf{98.75} \\ \hline

Gaussian& 58.43 / 39.41 / \textbf{0.0}   / 10.78 / \textbf{0.0}   / \textbf{0.0}   / \textbf{0.0}
        & 55.85 / 72.04 / \textbf{100.0} / 94.02 / \textbf{100.0} / \textbf{100.0} / \textbf{100.0} \\ \hline

Uniform & 32.04 / 18.49 / \textbf{0.0}   / 15.99 / \textbf{0.0}   / \textbf{0.0}   / \textbf{0.0}
        & 85.13 / 89.81 / \textbf{100.0} / 92.34 / \textbf{100.0} / \textbf{100.0} / \textbf{100.0} \\ \hline
        
\end{tabular}
\caption{Quantitative results on distinguishing test set of CIFAR-100 InD dataset from several OOD datasets. $\uparrow$ indicates larger value is better and $\downarrow$ indicates lower value is better. \textit{All values are percentages.}
\label{tab:cifar100}}
\end{table*}
\subsection{Details of the methods used in comparisons}
\label{sec:hyperparameter_selection}
In the classification experiments, we compare the proposed method with the Baseline method proposed by Hendryks et al. \cite{hendrycks17baseline}, ODIN \cite{liang2017enhancing}, Mahalanobis \cite{lee2018simple}, MCD \cite{yu2019unsupervised}, G-ODIN \cite{hsu2020generalized}, and EBM \cite{liu2020energy} which are primarily designed for OOD detection in classification tasks.

\vspace{0.1cm} \noindent ODIN and Mahalanobis have user defined parameters that significantly affect their performance. 
For ODIN, these parameters are temperature scaling and input pre-processing magnitude. For Mahalanobis, it is only the input pre-processing magnitude. In the original papers, these parameters are chosen from a list of values by utilizing a subset of the target OOD dataset as validation. Similarly, MCD requires samples from the target OOD dataset during training. However, in a real use case, we usually do not have access to the target OOD dataset. EBM uses a large-scale dataset that contains 80 Million images \cite{torralba200880} as OOD to fine-tune networks for OOD detection to obtain the best results reported in \cite{liu2020energy}. We argue that using such a large dataset as OOD in the training may not be suitable to evaluate the real performance of the method since the large dataset may contain examples very similar to the target OOD benchmarks used in the experiments. For a fair comparison of the existing methods, we use adversarial images for the methods, including ours, that require OOD examples either for setting the hyperparameters or model fine-tuning. Note that the adversarial images can be obtained with no additional cost using the InD training set. We searched the adversarial magnitude parameter from a set of values $\{0.01, 0.1, 1.0, 2.0, 5.0\}$ and choose the best values based on the performance on a hold-out set of adversarial images. The implementation of G-ODIN is not publicly available but as our experimental settings are very similar to theirs~\cite{hsu2020generalized}, including network architectures and OOD datasets, we directly use the results presented in their paper for comparison.

In the segmentation experiments, we compare the proposed method with 5 different OOD detection methods that can work on non-classification tasks: Baseline \cite{hendrycks17baseline}, Bishop \cite{bishop1994novelty}, SSL \cite{hendrycks2019using}, RaPP \cite{kim2020rapp}, and Multitask\_SSL \cite{venkatakrishnan2020self}. In the experiments with RaPP, we train an auto-encoder network on the InD dataset (COCO) by using a 10-layers encoder-decoder network architecture as suggested in the original paper~\cite{kim2020rapp}. SSL uses a self-supervised rotation network for OOD detection. To compare with SSL, we randomly rotate the training images of the InD dataset by a value from the set $\{0^\circ, 90^\circ, 180^\circ, 270^\circ\}$ and train a neural network to predict the rotation angles applied on input images using a ResNet architecture. In Multitask\_SSL, we constructed a 12-layers variational autoencoder (VAE) \cite{kingma2013auto}where an additional 2 dense layers are applied on top of the latent representation for rotation prediction as suggested in the original paper. We trained the architecture jointly for both tasks. In Baseline \cite{hendrycks17baseline}, we take the average of the softmax probabilities for the foreground classes as a confidence score for OOD detection, which is expected to be higher for InD and lower for OOD samples. Lastly, we also compare our method with Bishop's work~\cite{bishop1994novelty} to demonstrate the value of applying KDE on the lower-dimensional space rather than the high-dimensional input space.

% ======================================
% FIGURES CLASSIFICATION
% ======================================
\begin{figure*}[ht]
     \centering
     \begin{tabular}{ccccc}
     SVHN & TIN & LSUN & iSUN & Adversarial \\
     \includegraphics[scale = 0.2]{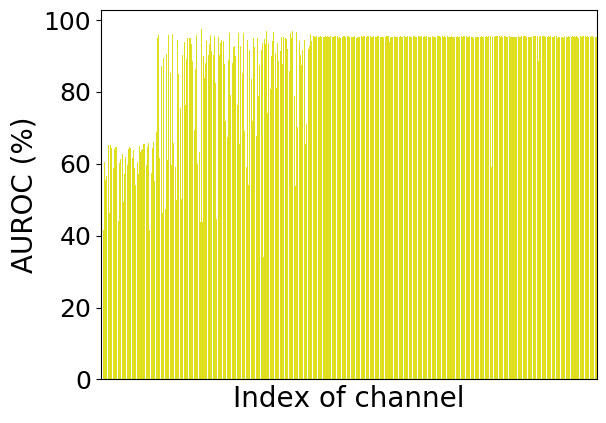} &
     \includegraphics[scale = 0.2]{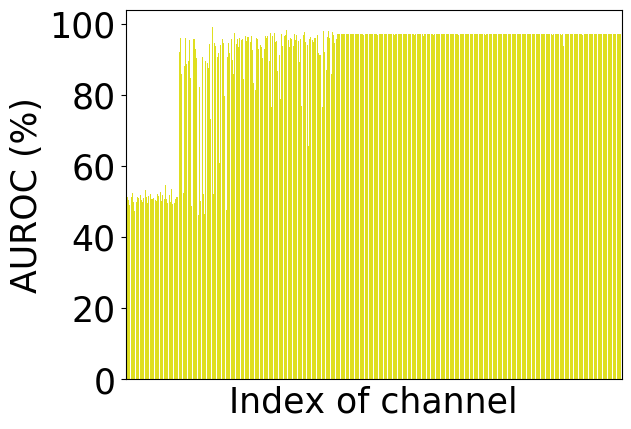} &
     \includegraphics[scale = 0.2]{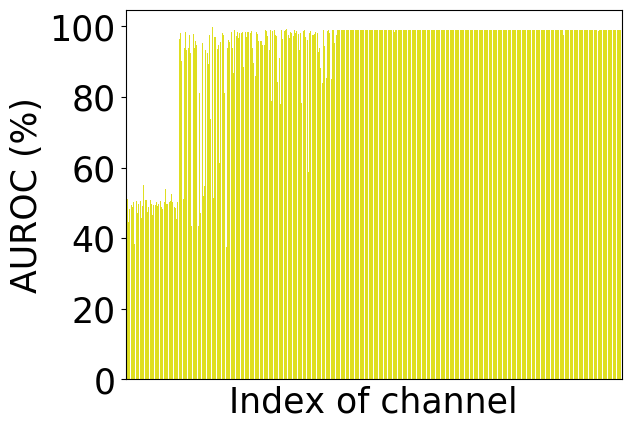} &
     \includegraphics[scale = 0.2]{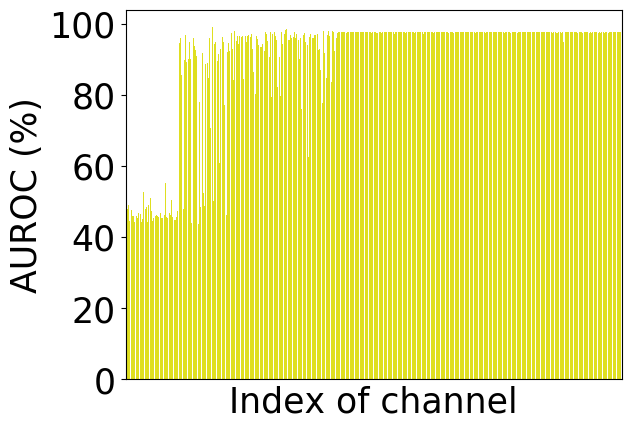} &
     \includegraphics[scale = 0.2]{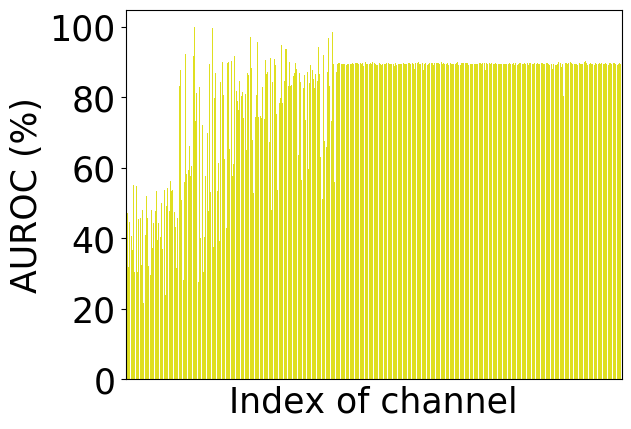}
     \end{tabular}
    \caption{Trends of channel-wise OOD detection performance evaluated with AUROC for different OOD datasets when the InD dataset is CIFAR-10. The plots show that the AUROC trend of adversarial images is similar to the OOD datasets, indicating that the logistic regression classifier trained using adversarial images can generalize to OOD ones.}
    \label{fig:AUROC_cifar10}
\end{figure*}
% ======================================
% EVALUATION METHODS
% ======================================
\subsection{Evaluation methods}
We use the evaluation methods that are commonly used for evaluating OOD detection methods in the literature \cite{hendrycks17baseline}. In all evaluations, we take the InD as the positive class and OOD as the negative class.

\vspace{0.1cm} \noindent\textbf{FPR at $95\%$ TPR:} False positive rate (FPR) is measured when the true positive rate (TPR) of $95\%$ is reached at a certain threshold. The false positive rate is calculated as FPR=FP/(FP+TN) and true positive rate is calculated as TPR=TP/(TP+FN) where TP, FP, TN, and FN represent true positive, false positive, true negative, and false negative, respectively.

\vspace{0.1cm} \noindent\textbf{AUROC:} The Area Under the Receiver Operating Characteristic curve (AUROC) is a threshold independent method that measures the area below Receiver Operating Characteristic (ROC) curve \cite{davis2006relationship}. ROC curve reflects the relationship between TPR and FPR values as the threshold changes. AUROC takes its highest value at $100\%$ when the detection is perfect.

\vspace{0.1cm} \noindent\textbf{Detection error:} This metric measures the probability of wrong classification when TPR is $95\%$. We denote the detection error by $P_{err}$ and compute as $P_{err}$=0.5(1-TPR+FPR). We present the results for this metric in the supplementary material for to save space.

% ======================================
% RESULTS AND ANALYSIS
% ======================================
\subsection{Results and Analysis}
Here we present the main quantitative results for both classification and segmentation tasks and present additional experimental results in the supplementary materials.

% ======================================
% RESULTS ON CLASSIFICATION NETWORKS
% ======================================
\subsubsection{Results on classification tasks}
We present the OOD detection results when CIFAR-10 and CIFAR-100 datasets are InD in Tables \ref{tab:cifar10} and \ref{tab:cifar100}, respectively. The results in CIFAR-10 experiments demonstrate that the proposed methods achieves better OOD detection performance than the existing methods on all OOD datasets. In the experiments on CIFAR-100 dataset, our method produces the best OOD detection results on all datasets except SVHN where it achieves the second best results. The results on the classification tasks suggest that the the proposed method improves the state-of-the-art OOD detection methods in almost all cases.

The logistic regression model in the proposed method is trained with the InD images and their adversarially perturbed versions as OOD samples. Even though the OOD space is vast, training with the perturbed samples seems to perform well. To better understand how the regression model generalizes to real OOD images, we analyzed the channel-wise confidence scores $\hat{p}_{lc}(x)$, which are the inputs to the regression model. We performed OOD detection with each channel's confidence score separately and plot the AUROC values for SVHN, TIN, LSUN, iSUN datasets, and the adversarial InD images when the InD is CIFAR-10 in Fig.~\ref{fig:AUROC_cifar10}. We do not present similar plots for Gaussian and Uniform datasets for the sake of brevity. It can be seen that the AUROC trends across channels for all OOD datasets are similar to the adversarial images, where AUROC results are lower in the earlier channels and higher in the later ones. This demonstrate that channel-wise scores of adversarial images generalize quite well to the ones obtained with the OOD images. This observation holds for CIFAR-100 dataset as well and we present similar plots in the supplementary material.

% ======================================
% TABLES SEGMENTATION
% ======================================
\begin{table*}[!]
\centering
\begin{tabular}{ccc}
\hline
OOD     &   FPR at 95\% TPR $\downarrow$&   AUROC $\uparrow$ \\ \cline{2-3}
~ & \multicolumn{2}{c}{Baseline / Bishop / SSL /  RaPP / Multitask\_SSL / \textbf{Proposed}} \\ \hline
ABIDE\_Stanford\_T1w    & 78.90 / 76.21 / 49.60 / 69.57 / 66.48 / \textbf{44.25}
                        & 48.26 / 81.71 / 63.45 / 52.44 / 54.30 / \textbf{89.27} \\ \hline
                        % & 41.95 / 40.60 / 27.30 / 37.28 / 35.74 / \textbf{24.62} \\ \hline
HCP\_T1w                & 88.06 / 79.72 / 63.51 / 87.20 / 76.58 / \textbf{42.93}
                        & 39.30 / 75.83 / 55.13 / 40.11 / 45.35 / \textbf{93.96} \\ \hline
                        % & 46.53 / 42.36 / 34.25 / 46.10 / 40.79 / \textbf{23.96} \\ \hline
HCP\_T2w                & 80.37 / 41.77 / 81.28 / 57.85 / 70.39 / \textbf{40.27}
                        & 42.19 / 92.93 / 43.82 / 52.78 / 47.88 / \textbf{94.62} \\ \hline
                        % & 42.68 / 23.38 / 43.14 / 31.42 / 37.69 / \textbf{22.63} \\ \hline
\end{tabular}
\caption{Quantitative results on distinguishing test set of ABIDE\_Caltech\_T1w InD dataset from several OOD datasets. $\uparrow$ indicates larger value is better and $\downarrow$ indicates lower value is better. \textit{All values are percentages.} \label{tab:abide_caltech_t1w}}
\end{table*}
\begin{table*}[!]
\centering
\begin{tabular}{ccc}
\hline
OOD     &   FPR at 95\% TPR $\downarrow$&   AUROC $\uparrow$ \\ \cline{2-3}
~ & \multicolumn{2}{c}{Baseline / Bishop / SSL /  RaPP / Multitask\_SSL /\textbf{Proposed}} \\ \hline
ABIDE\_Stanford\_T1w    & 59.25 / 100.0 / 62.96 / 67.01 / 45.06 / \textbf{44.78}
                        & 71.34 / 39.02 / 84.43 / 67.07 / 83.78 / \textbf{90.42} \\ \hline
                        % & 32.12 / 52.50 / 33.98 / 36.00 / \textbf{24.53} / 24.89 \\ \hline
ABIDE\_Caltech\_T1w     & 83.26 / 100.0 / 58.94 / 99.68 / 63.88 / \textbf{11.71}
                        & 68.41 / 17.21 / 87.22 / 59.38 / 79.56 / \textbf{96.77} \\ \hline
                        % & 44.43 / 52.50 / 31.97 / 52.34 / 34.44 / \textbf{8.35} \\ \hline
HCP\_T2w                & 47.55 / 94.98 / 76.89 / 47.94 / 61.62 / \textbf{18.77}
                        & 72.88 / 62.12 / 57.79 / 70.56 / 73.39 / \textbf{95.60} \\ \hline
                        % & 26.27 / 49.99 / 40.94 / 26.47 / 33.31 / \textbf{11.88} \\ \hline
\end{tabular}
\caption{Quantitative results on distinguishing test set of HCP\_T1w InD dataset from several OOD datasets. $\uparrow$ indicates larger value is better and $\downarrow$ indicates lower value is better. \textit{All values are percentages.} \label{tab:hcp_t1w}}
\end{table*}
% ======================================
% RESULTS ON SEGMENTATION NETWORKS
% ======================================
\subsubsection{Results on segmentation tasks}
\begin{figure}
    \centering
    \includegraphics[scale = 0.22]{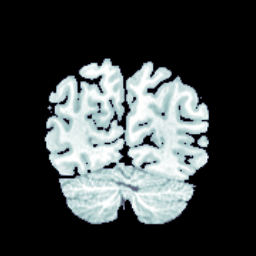}
    \includegraphics[scale = 0.22]{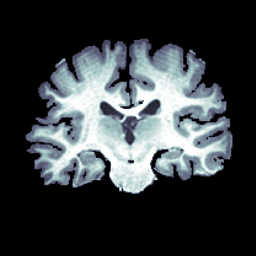}
    \includegraphics[scale = 0.22]{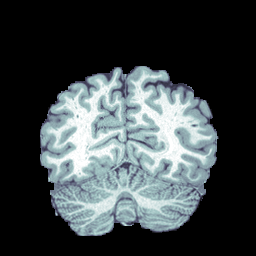}
    \includegraphics[scale = 0.22]{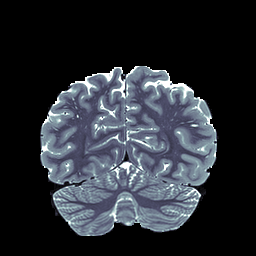} \\
    \includegraphics[scale = 0.22]{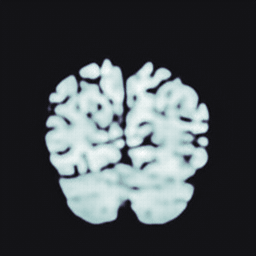}
    \includegraphics[scale = 0.22]{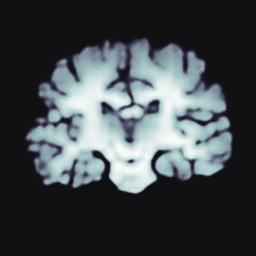}
    \includegraphics[scale = 0.22]{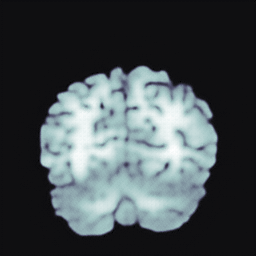}
    \includegraphics[scale = 0.22]{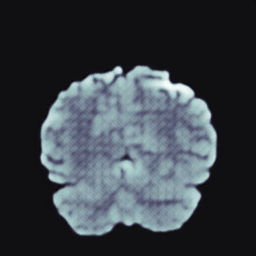}
    \caption{Input images (first row) and reconstructions (second row) of the autoencoder trained on ABIDE\_Caltech\_T1w dataset for RaPP. From left to right, the images belong to ABIDE\_Caltech\_T1w, ABIDE\_Stanford\_T1w, HCP\_T1w, and HCP\_T2w datasets, respectively.}
    \label{fig:visual_example2}
\end{figure}

We present the OOD detection results when InD datasets are ABIDE\_Caltech\_T1w and HCP\_T1w dataset in Tables \ref{tab:abide_caltech_t1w} and \ref{tab:hcp_t1w}, respectively. The results demonstrate that the proposed method improves the existing methods in all cases. Since Bishop \cite{bishop1994novelty} works on high-dimensional input space, it cannot achieve accurate density estimation and produces poor OOD detection results as expected. Here, the results of the self-supervised methods: SSL, RaPP, and Multitask\_SSL, were lower than we expect, and we investigated further to interpret the results better. These methods exhibit diminished performance because the self-supervised networks generalize surprisingly well to OOD images. For example, the network trained on HCP\_T1w images for the SSL rotation task predicts the rotation angles with $\approx75\%$ accuracy for both InD and OOD datasets. This holds for the case when we use ABIDE\_Caltech\_T1w as InD. Our observation is similar to the autoencoder network trained for RaPP. In Figure \ref{fig:visual_example2}, we show input images and their reconstructions when the ABIDE\_Caltech\_T1w is InD. The visual results show that the autoencoder reconstructs images from different datasets equally well.

% ======================================
% CHANNEL-WISE VS LAYER-WISE
% ======================================
\begin{table}[]
\begin{tabular}{ccc|cc} \hline
\multirow{2}{*}{} & \multicolumn{2}{c|}{Layer-wise}                                           & \multicolumn{2}{c}{Channel-wise}                                  \\ \cline{2-5}
                  & \begin{tabular}[c]{@{}c@{}}Gaussian\\ (Mahalanobis)\end{tabular} & KDE   & Gaussian & \begin{tabular}[c]{@{}c@{}}KDE\\ (Proposed)\end{tabular} \\ \hline
SVHN    & 15.53 & 24.09 & 12.00 & 17.46  \\ \hline
TIN     & 24.33 & 34.08 & 13.90 & 7.64 \\ \hline
LSUN    & 28.68 & 28.47 & 5.10 & 3.73 \\ \hline
iSUN    & 29.46 & 33.49 & 7.50 & 6.07 \\ \hline 
\end{tabular}
\caption{Comparison between different combinations of density estimation methods (Gaussian and KDE) with feature spaces (layer-wise and channel-wise) in terms of FPR at 95\% TPR in CIFAR-100 dataset. \label{tab:combinations}}
\end{table}
\subsection{Channel-wise vs layer-wise and KDE vs parametric estimation} 
In the proposed method, we perform channel-wise KDE. Compared to the closest work Mahalanobis~\cite{lee2018simple}, this introduces two changes in the density estimation, one in feature selection (layer-wise vs channel features) and the other in estimation methodology (KDE vs Gaussian). In this section, we quantify the contribution of each change.
To this end, we perform OOD detection with all possible combinations. The results in Table \ref{tab:combinations} demonstrate that performing channel-wise density estimation leads to a large improvement on OOD detection accuracy compared to layer-wise density estimation. We argue that this improvement is due to achieving more accurate density estimation in 1D space with the channel-wise features. Dependencies between channels are taken into account in the logistic regression model. We also observe that performing KDE on the channel-wise features yields further improvements over using Gaussian in most cases. This is expected since KDE is more flexible and can lead to more accurate density estimations. We present further experiments to compare channel-wise vs layer-wise KDE in the supplementary material.

% ======================================
% COMPUTATION TIME
% ======================================
\subsection{Computation time}
\begin{figure}
    \centering
    \includegraphics[scale = 0.35]{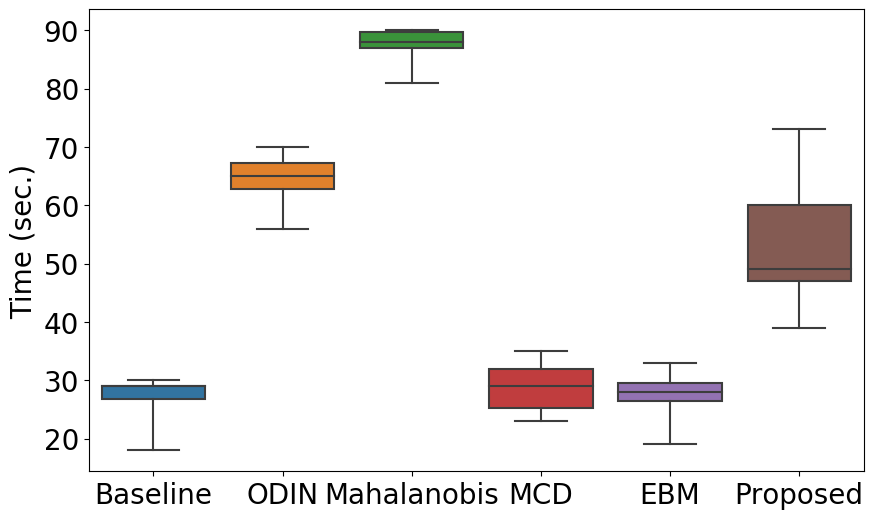}
    \caption{Mean and standard deviation of the computation times of all OOD datasets in the CIFAR-10 experiments.}
    \label{fig:computation_time}
\end{figure}

We compare the proposed method with the others in terms of computation time. We measure the computation time needed to perform detection on all OOD datasets in the CIFAR-10 experiments for Baseline, ODIN, MCD, EBM, and Mahalanobis. Computation time of G-ODIN is not presented since the implementation is not publicly available. We show the mean and standard deviation of the computation times in the plot in Fig. \ref{fig:computation_time}. The results demonstrate that the Baseline, MCD, and EBM are the fastest methods as expected since they do only a single forward pass in the networks. The proposed method comes after them despite the computational cost of KDE. ODIN and Mahalanobis are the computationally less efficient than the other due to the input pre-processing step, which is computationally costly. Mahalanobis is the slowest method among all since it applies input pre-processing before computing the scores at each layer.

\section{Conclusion}
In this paper, we presented a task-agnostic OOD detection method that estimates feature densities for each channel of a DNN using KDE. Features of a test image are evaluated at the corresponding KDEs to obtain a confidence score per channel, which is expected to be higher for InD images than OOD ones. These scores are combined into a final score using logistic regression classifier, that is pre-trained using InD training images and their adversarially perturbed versions. Being task-agnostic, the proposed method can be applied to both classification and non-classification DNNs. 

\vspace{0.1cm} \noindent We performed an extensive evaluations on both classification and segmentation networks and compare them with the state-of-the-art methods. The results demonstrate that the proposed method that uses channel-wise KDE improves state-of-the-art in majority of the cases. We also performed an experiment to compare the channel-wise density estimation with layer-wise estimation and performing parametric density estimation (Gaussian) with the nonparametric one (KDE). The results show that channel-wise density estimation plays the major role in improving the results since density estimation is easier lower dimensional space. Using KDE for density estimation instead of assuming that the pdf is Gaussian leads to further improvement thanks to the flexibility of the nonparametric method.
% The results demonstrate that the proposed method attains high OOD detection accuracy across different tasks, exhibiting a larger scope of applications than existing task-specific methods and improving state-of-the-art for task-agnostic methods.
\appendix

\section{Results with \emph{Detection Error} metric}
The experimental results in the main paper are presented with respect to FPR at 95\% TPR and AUROC metrics. In addition to these results, we present results with respect to Detection Error metric for all methods in Figures \ref{tab:cifar10_detection_error}, \ref{tab:cifar100_detection_error}, \ref{tab:abide_caltech_t1w_detection_error}, and \ref{tab:hcp_t1w_detection_error} for CIFAR-10, CIFAR-100, ABIDE\_Caltech\_T1w, and HCP\_T1w dataset, respectively.
\begin{table*}
\centering
\begin{tabular}{cc}
\hline
OOD     &   Detection Error $\downarrow$ \\ \cline{2-2}
~ & Baseline / ODIN / Mahalanobis / MCD / EBM / \textbf{Proposed} \\ \hline
SVHN    & 15.38 / 11.32 / 5.90 / 32.80 / 12.45 / \textbf{5.74} \\ \hline
        
TIN     & 16.68 / 8.12 / 9.44 / 22.72 / 27.76 / \textbf{6.70} \\ \hline

LSUN    & 16.65 / 7.69 / 9.02 / 19.73 / 22.10 / \textbf{4.40} \\ \hline

iSUN    & 16.51 / 8.68 / 10.23 / 21.36 / 22.99 / \textbf{6.15} \\ \hline

Gaussian& 5.72 / 3.84 / \textbf{0.0} / 4.60 / 2.50 / \textbf{0.0} \\ \hline

Uniform & 7.12 / 4.58 / \textbf{0.0} / 9.08 / 2.50 / \textbf{0.0} \\ \hline
\end{tabular}
\caption{\emph{Detection Error} metric results on distinguishing test set of CIFAR-10 InD dataset from several OOD datasets. \textit{All values are percentages.}
\label{tab:cifar10_detection_error}}
\end{table*}
\begin{table*}
\centering
\begin{tabular}{cc}
\hline
OOD     &   Detection Error $\downarrow$ \\ \cline{2-2}
~ & Baseline / ODIN / Mahalanobis / MCD / EBM / \textbf{Proposed} \\ \hline
SVHN    & 30.36 / 14.88 / \textbf{7.72} / 36.16 / 19.64 / 11.23 \\ \hline
        
TIN     & 31.98 / 19.37 / 11.71 / 30.94 / 35.85 /\textbf{6.32} \\ \hline

LSUN    & 34.85 / 21.04 / 12.39 / 31.70 / 33.62 /\textbf{4.36} \\ \hline

iSUN    & 34.13 / 21.60 / 13.37 / 34.66 / 34.77 /\textbf{5.53} \\ \hline

Gaussian& 31.71 / 22.20 / \textbf{0.0} / 7.89 / \textbf{0.0} /\textbf{0.0} \\ \hline

Uniform & 18.52 / 11.74 / \textbf{0.0} / 10.49 / \textbf{0.0} /\textbf{0.0} \\ \hline
\end{tabular}
\caption{\emph{Detection Error} metric results on distinguishing test set of CIFAR-100 InD dataset from several OOD datasets. \textit{All values are percentages.}
\label{tab:cifar100_detection_error}}
\end{table*}

\begin{table*}[!]
\centering
\begin{tabular}{cc}
\hline
OOD     &   Detection Error $\downarrow$ \\ \cline{2-2}
~ & Baseline / Bishop / SSL /  RaPP / Multitask\_SSL / \textbf{Proposed} \\ \hline
ABIDE\_Stanford\_T1w    & 41.95 / 40.60 / 27.30 / 37.28 / 35.74 / \textbf{24.62} \\ \hline
HCP\_T1w                & 46.53 / 42.36 / 34.25 / 46.10 / 40.79 / \textbf{23.96} \\ \hline
HCP\_T2w                & 42.68 / 23.38 / 43.14 / 31.42 / 37.69 / \textbf{22.63} \\ \hline
\end{tabular}
\caption{\emph{Detection Error} metric results on distinguishing test set of ABIDE\_Caltech\_T1w InD dataset from several OOD datasets. \textit{All values are percentages.} \label{tab:abide_caltech_t1w_detection_error}}
\end{table*}

\begin{table*}[!]
\centering
\begin{tabular}{cc}
\hline
OOD     &  Detection Error $\downarrow$ \\ \cline{2-2}
~ & Baseline / Bishop / SSL /  RaPP / Multitask\_SSL /\textbf{Proposed} \\ \hline
ABIDE\_Stanford\_T1w    & 32.12 / 52.50 / 33.98 / 36.00 / \textbf{24.53} / 24.89 \\ \hline
ABIDE\_Caltech\_T1w     & 44.43 / 52.50 / 31.97 / 52.34 / 34.44 / \textbf{8.35} \\ \hline
HCP\_T2w                & 26.27 / 49.99 / 40.94 / 26.47 / 33.31 / \textbf{11.88} \\ \hline
\end{tabular}
\caption{\emph{Detection Error} metric results on distinguishing test set of HCP\_T1w InD dataset from several OOD datasets. \textit{All values are percentages.} \label{tab:hcp_t1w_detection_error}}
\end{table*}

\section{Results with different number of samples $N$ in KDE}
$N$ is a hyperparameter of the proposed method that determines number of samples to be used for unbiased estimation of the target density using KDE. In this section, we analyze the behavior of the proposed approach as a function of different $N$ values. We present the OOD detection performance using different metrics for both CIFAR-10 and CIFAR-100 experiments in Tables \ref{tab:N_values_cifar10} and \ref{tab:N_values_cifar100}, respectively. There results demonstrate that the proposed method is not very sensitive to the number of samples, $N$, used in KDE.
\begin{table*}[ht]
\centering
\begin{tabular}{ccccc}
\hline
OOD     &   FPR at 95\% TPR $\downarrow$    &   AUROC $\uparrow$ & Detection Error $\downarrow$ \\ \cline{2-4}
~ & \multicolumn{3}{c}{$N =$ 1000/2000/5000/7000} \\ \hline
SVHN        & 23.16 / 19.33 / 6.49 / 17.22  & 94.19 / 94.57 / 98.48 / 97.89 & 14.08 / 12.16 / 5.74 / 11.11  \\ \hline
TIN         & 7.02 / 8.45 / 8.41 / 8.02     & 98.29 / 98.26 / 98.31 / 98.73 & 6.01 / 6.72 / 6.70 / 6.51     \\ \hline
LSUN        & 1.92 / 2.40 / 3.80 / 1.84     & 99.36 / 99.29 / 99.01 / 99.46 & 3.46 / 3.70 / 4.40 / 3.42     \\ \hline
iSUN        & 5.63 / 6.69 / 7.31 / 5.91     & 98.62 / 98.63 / 98.55 / 98.85 & 5.31 / 5.84 / 6.15 / 5.45     \\ \hline
Gaussian    & 0.0 / 0.0 / 0.0 / 0.0         & 100.0 / 100.0 / 100.0 / 100.0 & 0.0 / 0.0 / 0.0 / 0.0         \\ \hline
Uniform     & 0.0 / 0.0 / 0.0 / 0.0         & 100.0 / 100.0 / 100.0 / 100.0 & 0.0 / 0.0 / 0.0 / 0.0         \\ \hline
\end{tabular}
\caption{Performance of the proposed method on CIFAR-10 dataset as a function of number of samples ($N$) used in KDE.
\label{tab:N_values_cifar10}}
\end{table*}
\begin{table*}[ht]
\centering
\begin{tabular}{ccccc}
\hline
OOD     &   FPR at 95\% TPR $\downarrow$    &   AUROC $\uparrow$ & Detection Error $\downarrow$ \\ \cline{2-4}
~ & \multicolumn{3}{c}{$N =$ 1000/2000/5000/7000} \\ \hline
SVHN        & 18.65 / 17.19 / 17.46 / 17.69 & 95.61 / 96.24 / 95.44 / 95.99 & 11.82 / 11.09 / 11.23 / 11.34 \\ \hline
TIN         & 14.98 / 11.98 / 7.64 / 10.96  & 96.81 / 97.72 / 98.38 / 98.01 & 9.99 / 8.49 / 6.32 / 7.98     \\ \hline
LSUN        & 10.69 / 6.65 / 3.73 / 5.17    & 97.60 / 98.65 / 99.13 / 98.98 & 7.84 / 5.82 / 4.36 / 5.08     \\ \hline
iSUN        & 12.41 / 9.01 / 6.07 / 7.21    & 97.33 / 98.24 / 98.75 / 98.62 & 8.70 / 7.00 / 5.53 / 6.10     \\ \hline
Gaussian    & 0.0 / 0.0 / 0.0 / 0.0         & 100.0 / 100.0 / 100.0 / 100.0 & 0.0 / 0.0 / 0.0 / 0.0         \\ \hline
Uniform     & 0.0 / 0.0 / 0.0 / 0.0         & 100.0 / 100.0 / 100.0 / 100.0 & 0.0 / 0.0 / 0.0 / 0.0         \\ \hline
\end{tabular}
\caption{Performance of the proposed method on CIFAR-100 dataset as a function of number of samples ($N$) used in KDE.
\label{tab:N_values_cifar100}}
\end{table*}

\section{Details of the adversarial perturbation method - FGSM}
As we discussed in Sec. 2.2. of the main paper, we produce OOD samples to train logistic regression using an adversarial attack method called Fast Gradient Sign Method (FGSM) \cite{goodfellow2014explaining}. In this section, we present more details of FGSM.

\begin{figure}
    \centering
    \includegraphics[scale = 2.0]{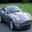}
    \includegraphics[scale = 2.0]{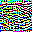}
    \includegraphics[scale = 2.0]{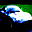} \\
    \includegraphics[scale = 2.0]{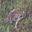}
    \includegraphics[scale = 2.0]{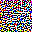}
    \includegraphics[scale = 2.0]{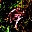}
    \caption{Applying FGSM to some example images from CIFAR-10 dataset. Note that each row corresponds to a different image. Columns from left to right indicate original image, gradient of the loss w.r.t. input image, and adversarial image, respectively.}
    \label{fig:FGSM}
\end{figure}

The FGSM method uses gradients of the neural network when creating an adversarial example. To this end, it computes the gradient of the loss function with respect to the input image and adds the gradient to the original input image to obtain the adversarial image. Perturbing the input image with the gradient leads to maximizing the loss function. Adversarial perturbation is applied to an input image $x$ using the following equation
\[
x_{adv} = x + \epsilon \times sign(\nabla(L(x, y)))
\]
where $y$ is the label of $x$, $L(x,y)$ is the loss function, and $\epsilon$ is a parameter to control the strength of the perturbation. We present some example images from CIFAR-10 dataset along with their adversarial version in Fig. \ref{fig:FGSM}.

\section{Hyperparameter search for $\epsilon$ in FGSM}
In our experiments, we use adversarial examples generated by FGSM when applying ODIN, Mahalanobis, MCD, EBM, and the proposed method since they require some OOD samples either for hyperparameter selection or training. As mentioned in the main paper, we selected the best $\epsilon$ value from a set of values $\{0.01, 0.1, 1.0, 2.0, 5.0\}$ in our experiments based on the performance on a validation set which is a portion of the adversarial examples. In this section, we present the results obtained with each $\epsilon$ for Mahalanobis (Table \ref{tab:Mahalanobis_epsilon}), EBM (Table \ref{tab:EBM_epsilon}), and the proposed method (Table \ref{tab:proposed_epsilon}) to show the sensitivity to this parameter.
\begin{table*}[ht]
\centering
\begin{tabular}{cccccc}
\hline
\multicolumn{6}{c}{Mahalanobis} \\ \hline
$\epsilon = $   & 0.01 & 0.1 & 1.0 & 2.0 & 5.0 \\ \hline         
            &   \multicolumn{5}{c}{ CIFAR-10 / CIFAR-100} \\ \hline
SVHN        & 44.25 / 32.39 & 26.46 / 19.72 & 8.37 / 15.53 & 4.33 / 14.84 & 4.70 / 37.36 \\ \hline
TIN         & 12.00 / 24.73 & 10.07 / 22.66 & 18.89 / 24.33 & 16.09 / 26.55 & 11.57 / 25.90 \\ \hline
LSUN        & 10.72 / 25.42 & 8.70 / 27.43 & 19.61 / 28.68 & 15.98 / 32.30 & 11.05 / 25.85 \\ \hline
iSUN        & 14.33 / 28.48 & 11.87 / 28.55 & 22.46 / 29.46 & 18.22 / 32.16 & 13.27 / 29.88 \\ \hline
Gaussian    & 0.0 / 0.0 & 0.0 / 0.0 & 0.0 / 0.0 & 0.0 / 0.0 & 0.0 / 0.0 \\ \hline
Uniform     & 0.0 / 0.0 & 0.0 / 0.0 & 0.0 / 0.0 & 0.0 / 0.0 & 0.0 / 0.0 \\ \hline
\end{tabular}
\caption{Results of Mahalanobis in terms of FPR at 95\% TPR on CIFAR-10 and CIFAR-100 for different $\epsilon$ values.
\label{tab:Mahalanobis_epsilon}}
\end{table*}
\begin{table*}[ht]
\centering
\begin{tabular}{cccccc}
\hline
\multicolumn{6}{c}{EBM} \\ \hline
$\epsilon = $   & 0.01 & 0.1 & 1.0 & 2.0 & 5.0 \\ \hline         
            &   \multicolumn{5}{c}{ CIFAR-10 / CIFAR-100} \\ \hline
SVHN        & 0.03 / 1.39 & 4.21 / 18.35 & 6.86 / 45.49 & 77.59 / 89.0 & 57.83 / 92.94 \\ \hline
TIN         & 76.13 / 87.07 & 61.68 / 80.92 & 35.88 / 70.04 & 30.40 / 68.86 & 37.41 / 70.91 \\ \hline
LSUN        & 67.18 / 89.12 & 44.95 / 76.59 & 21.62 / 67.99 & 21.10 / 64.93 & 28.63 / 67.77 \\ \hline
iSUN        & 66.54 / 87.04 & 46.88 / 76.76 & 22.52 / 70.11 & 22.0 / 65.93 & 28.43 / 69.89 \\ \hline
Gaussian    & 99.75 / 99.44 & 91.19 / 93.42 & 0.13 / 0.0 & 0.0 / 0.0 & 0.0 / 0.0 \\ \hline
Uniform     & 80.13 / 100.0 & 53.87 / 94.59 & 0.0 / 0.0 & 0.0 / 0.0 & 0.0 / 0.0 \\ \hline
\end{tabular}
\caption{Results of EBM in terms of FPR at 95\% TPR on CIFAR-10 and CIFAR-100 for different $\epsilon$ values.
\label{tab:EBM_epsilon}}
\end{table*}

\begin{table*}[ht]
\centering
\begin{tabular}{cccccc}
\hline
\multicolumn{6}{c}{Proposed} \\ \hline
$\epsilon = $   & 0.01 & 0.1 & 1.0 & 2.0 & 5.0 \\ \hline         
            &   \multicolumn{5}{c}{ CIFAR-10 / CIFAR-100} \\ \hline
SVHN        & 17.54 / 27.71 & 12.03 / 28.24 & 6.49 / 17.46 & 7.18 / 18.79 & 4.08 / 43.36 \\ \hline
TIN         & 32.46 / 52.03 & 17.27 / 25.17 & 8.41 / 7.64 & 10.33 / 13.73 & 19.03 / 31.05 \\ \hline
LSUN        & 21.44 / 49.67 & 11.82 / 17.55 & 3.80 / 3.73 & 4.62 / 8.60 & 5.30 / 16.29 \\ \hline
iSUN        & 26.83 / 52.90 & 14.92 / 17.80 & 7.31 / 6.07 & 8.29 / 12.25 & 13.72 / 31.11 \\ \hline
Gaussian    & 3.04 / 0.0 & 0.0 / 0.0 & 0.0 / 0.0 & 0.0 / 0.0 & 0.0 / 0.0  \\ \hline
Uniform     & 1.17 / 0.0 & 0.0 / 0.0 & 0.0 / 0.0 & 0.0 / 0.0 & 0.0 / 0.0  \\ \hline
\end{tabular}
\caption{Results of the proposed method in terms of FPR at 95\% TPR on CIFAR-10 and CIFAR-100 for different $\epsilon$ values.
\label{tab:proposed_epsilon}}
\end{table*}

\section{AUROC trends for CIFAR-100 datasets}
\begin{figure*}
\centering
\includegraphics[scale = 0.2]{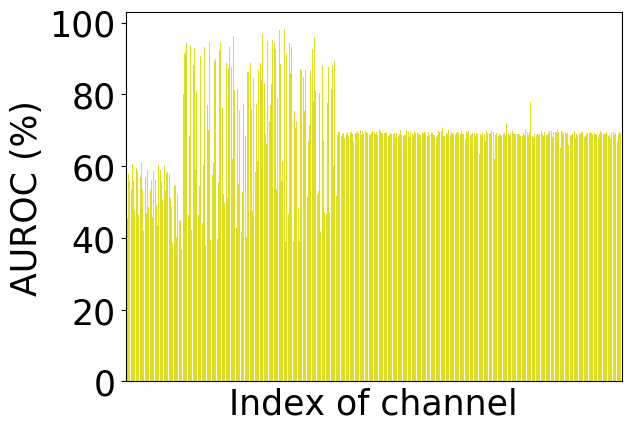}
\includegraphics[scale = 0.2]{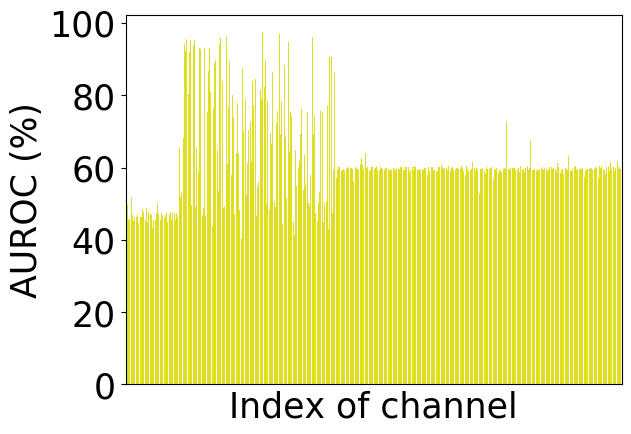}
\includegraphics[scale = 0.2]{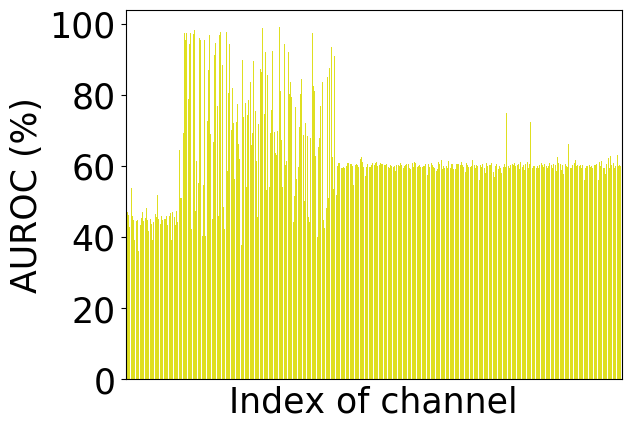}
\includegraphics[scale = 0.2]{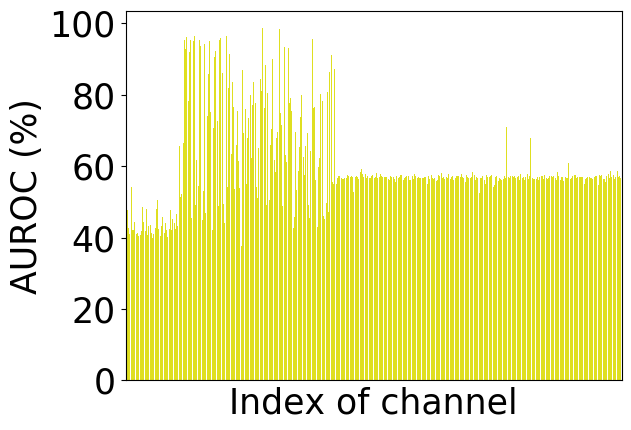}
\includegraphics[scale = 0.2]{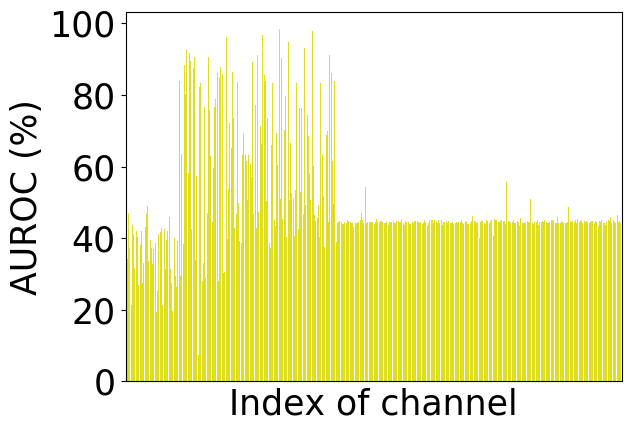}
\caption{Trends of channel-wise OOD detection performance evaluated with AUROC for different OOD datasets when the InD dataset is CIFAR-100. The plots show that the AUROC trend of adversarial images is similar to most of the OOD datasets, indicating that the logistic regression classifier trained using adversarial images can generalize to most of the OOD datasets.}
\label{fig:AUROC_cifar100}
\end{figure*}

In the main paper, we perform analysis on CIFAR-10 dataset to better understand how the regression model trained on adversarial images generalizes to real OOD images. In this section, we perform similar analysis on CIFAR-100 to demonstrate that the analysis on CIFAR-10 holds for the other datasets. We performed OOD detection with each channel's confidence score separately and plot the AUROC values for SVHN, TIN, LSUN, iSUN datasets, and the adversarial InD images when the InD is CIFAR-100 in Fig.~\ref{fig:AUROC_cifar100}. We observe that AUROC trends across channels for all OOD datasets are similar to the adversarial images. This demonstrate that channel-wise scores of adversarial images generalize quite well to the ones obtained with the OOD images.

\section{Comparison between channel-wise vs Layer-wise KDE}
In this section, we present additional results to show the improvement that is achieved by using channel-wise density estimation instead of layer-wise on CIFAR-10 (Table \ref{tab:cifar10_layer_channel_wise}) and CIFAR-100 (Table \ref{tab:cifar100_layer_channel_wise}) datasets. Although, layer-wise KDE achieves slightly better results on CIFAR-10 dataset, channel-wise KDE performs significantly better on CIFAR-100.
\begin{table*}[ht]
\centering
\begin{tabular}{cccc}
\hline
OOD     &   FPR at 95\% TPR $\downarrow$ & AUROC $\uparrow$ & Detection Error $\downarrow$\\ \cline{1-4}
~ & \multicolumn{3}{c}{Layer-wise KDE / \textbf{Proposed} (Channel-wise KDE)} \\ \hline
SVHN    &  2.44 / 6.49
        &  99.36 / 98.48
        &  3.72 / 5.74 \\ \hline
TIN     &  5.11 / 8.41
        &  98.99 / 98.31
        &  5.05 / 6.70 \\ \hline
LSUN    &  2.54 / 3.80
        &  99.44 / 99.01
        &  3.77 / 4.40 \\ \hline
iSUN    &  3.83 / 7.31
        &  99.26 / 98.55
        &  4.41 / 6.15 \\ \hline
Gaussian & {0.0} / {0.0} 
         & {100.0} / {100.0}
         & {0.0} / {0.0} \\ \hline
Uniform  & {0.0} / {0.0} 
         & {100.0} / {100.0}
         & {0.0} / {0.0} \\ \hline
\end{tabular}
\caption{Comparison between channel-wise vs layer-wise KDE on CIFAR-10 dataset. \label{tab:cifar10_layer_channel_wise}}
\end{table*}
\begin{table*}[h]
\centering
\begin{tabular}{cccc}
\hline
OOD     &   FPR at 95\% TPR $\downarrow$ & AUROC $\uparrow$ & Detection Error $\downarrow$\\ \cline{1-4}
~ & \multicolumn{3}{c}{Layer-wise KDE / \textbf{Proposed} (Channel-wise KDE)} \\ \hline
SVHN    & 24.09 / 17.46 
        & 94.66 / 95.44
        & 14.54 / 11.23 \\ \hline

TIN     & 34.08 / 7.64
        & 93.89 / 98.38
        & 19.54 / 6.32 \\ \hline

LSUN    & 28.47 / 3.73
        & 94.55 / 99.13 
        & 16.73 / 4.36 \\ \hline

iSUN    & 33.49 / 6.07
        & 93.90 / 98.75 
        & 19.24 / 5.53 \\ \hline
        
Gaussian & {0.0} / {0.0} 
         & {100.0} / {100.0}
         & {0.0} / {0.0} \\ \hline
Uniform  & {0.0} / {0.0} 
         & {100.0} / {100.0}
         & {0.0} / {0.0} \\ \hline
\end{tabular}
\caption{Comparison between channel-wise vs layer-wise KDE on CIFAR-100 dataset. \label{tab:cifar100_layer_channel_wise}}
\end{table*}

\clearpage
\newpage
{
\bibliographystyle{ieee_fullname}
\bibliography{egbib}
}

\end{document}